# Who will accept my request? Predicting response of link initiation in two-way relation networks

Amin Javari*, Mehrab Norouzitabllab†, Mahdi Jalili‡

*University of Illinois at Urbana-Champaign     †Tehran University     ‡Royal Melbourne Institute of Technology

javari2@illinois.edu     m.norouzitallab@gmail.com     mahdi.jalili@rmit.edu.au

**Abstract**

Popularity of social networks has rapidly increased over the past few years, and daily lives interrupt without their proper functioning. Social networking platform provide multiple interaction types between individuals, such as creating and joining groups, sending and receiving messages, sharing interests and creating friendship relationships. This paper addresses an important problem in social networks analysis and mining that is how to predict link initiation feedback in two-way networks. Relationships between two individuals in a two-way network include a link invitation from one of the individuals, which will be an established link if it is accepted by the invitee. We consider a sport gaming social networking platform and construct a multilayer social network between a number of users. The network formed by the link initiation process is on one of the layers, while the other two layers include a messaging relationships and interactions between the users. We propose a methodology to solve the link initiation feedback prediction problem in this multilayer fashion. The proposed method is based on features extracted from meta-paths, i.e. paths defined between different individuals from multiples layers in multilayer networks. We proposed a cluster-based approach to handle the sparsity issue in the dataset. Experimental results show that the proposed method can provide accurate prediction that outperforms state-of-the-art methods.

## 1. Introduction

Users often make diverse connections on various social media applications [1, 2]. Social network analysis and mining has received much attention in recent years [3], which is mainly due to availability of large-scale datasets from various online social networks, and advances in computational infrastructures. Most of the studies on network modeling of social interactions have mainly considered one-way relations. In a one-way relational network, a connection between two nodes can be created when one of them activates the relation [4]. Message interaction in Facebook is an example of one-way relation, where a user sends a message to another one, without needing confirmation of the recipient user. In two-way relational networks on the other hand, a relation between two users' needs confirmation of both. In this type of relations, one side of the connection, often called *initiator* (or *invitor*), initiates the connection and the other side, called *recipient* (or *invitee*), accepts or rejects the link initiation. A *link* is formed only when the connection request is accepted by the invitee. In the Facebook example, to create a friendship relation between two users, one of them should send the request and the other one should confirm it, after which a friendship relation is created between them. In general, relations which are created based on a request and need confirmation of both sides of the relation can be classified as *two-way relations*. Although this type of relations is widespread in various environments, to the best of our knowledge, it has not yet been investigated in previous works.

One-way and two-ways networks are different from each other in various aspects. In a network of one-way relations, there is only a single type of connection between the users, whereas link creation in two-way networks involves positive and negative responses by the invitee. Let us denote a link receiving a positive response from the invitee as complete, and the one receiving a negative response as incomplete. Apart from these structural differences, the concept of a complete two-way relation differs from a one-way relation. A complete two-way relation implies commitment of both sides in the relation, while a one-way relation represents a connection activated by only one of the sides. Considering these differences, a number of research questions arise in networks of two-way relations, such as the extraction of evolutional patterns, the effect of negative responses in the creation of new link initiations, and the prediction of the response to a link initiation. In this paper, we specifically focus on the last problem that is to predict whether the recipient node accepts or rejects the link initiated by the initiator node. Without losing generality, in this paper we investigate networks with a single type of requests and binary class of responses. The methodology can be easily extended to networks with multiple types of requests and/or multi-class responses.

The problem of response prediction can be viewed as a problem related to reciprocal prediction. Given a directed homogeneous network (e.g. a messaging network), a relation between two users $u$ and $v$ is symmetric or reciprocated if they interact as peers (i.e. send messages to each other). On the other hand, when user $u$ sends a message to user $w$ and does not receive anything back, it is an unreciprocated relation from $u$ to $v$. In the reciprocal prediction, the problem is that whether the edge from $u$ to $v$ exists, given that the edge from $v$ to $u$ presents. In general, the reciprocal prediction problem can be modeled as a special case of the response prediction problem with single type of request and binary classes of responses. If the relation between $u$ and $v$ is reciprocal, it can be modeled as a request from $u$ to $v$ with a positive response, while an unreciprocated relation from $u$ to $v$ can be regarded as a request with a negative response.

Predicting the reaction of users to link initiation would have potential applications in various disciplines such as recommender systems [5-7]. The positive/negative response to a link initiation from a specific user can be predicted by analyzing the connectivity structure of the network. If the users are informed about the prediction results, they can better decide whether or not to send the friendship invitation. Indeed, a priori providing the likelihood of acceptance/rejection of a particular link initiation to the initiating user, will help the user to make informed decisions. In order to solve the response prediction problem, we model the link request as a directed edge from the initiator towards the recipient, and the response given to the link initiation as the sign of the edge, positive for acceptance and negative for rejection. Having this model, the problem of response prediction is similar to the problem of edge sign prediction in signed networks [8-10]. Networks with both positive and negative signs have been recently studied in various works [9, 11-14]. However, what we pursue here is different from the previous studies of signed networks in two main aspects. First, the previous works used a signed edge from $u$ to $v$ to represent the attitude of $u$ towards $v$ – positive relation indicating friendship, trust or support and negative link to express disagreement, enmity or distrust. However, we employ signed networks to represent a network of two-way relations: positive and negative edges indicating complete and incomplete two-way relations, respectively. Clearly, considering the difference in the meaning of the edges, the prediction models proposed for sign prediction will not necessarily work well on the signed networks obtained from two-way relations.

We apply our proposed link initiation feedback prediction framework on data gathered from a social gaming website (mastercup.varzesh3.com). The dataset consists of a set of two-way relations among the users. Moreover, it contains two other types of relations, which can be considered as one-way relations. Indeed, the connection structure between users in this dataset is a multi-layer network [15], in which one layer is a network of two-way relations and the others are networks of one-way relations. Our aim here is to use such a multilayer structure of the connections, and predict the signs in the two-way relation layer (i.e., feedback to the link initiation), given the structure of other layers. To build the prediction model, we define the topological features based on paths between the target nodes. In order to extract the features, we introduce a method called cluster-based meta-paths. We show that the features obtained based on cluster-based meta-paths are more robust against sparsity of the input network compared with those extracted using classic meta-paths. The experimental results reveal effectiveness of the proposed model in link initiation response prediction.

## 2. Related works

### 2.1. Sign prediction in homogeneous networks

In the following, we discuss two state-of-the-art models introduced for predicting the signs in homogeneous (i.e., single layer) networks [16]. Sign prediction methods using path-based features have been studied in a number of works [17-19]. In general, these methods have two steps. First, a set of path-based features are extracted, and then a prediction model is built based on a machine learning algorithm. For instance, Leskovec et al. introduced a set of features based on social and status theories (combined with a set of degree-based features) to predict the signs [11]. Chiang et al. extended this model and showed that defining features based on longer paths between the target nodes can improve the prediction accuracy [20]. Shahriari et al. proposed node-based features, called reputation and optimism, and showed one can obtain much better results by taking such features in the learning process [21]. Shahriari and Jalili used node ranking strategies for link sign prediction and showed they result in rich information for the prediction process [22]. Khodadaddi and Jalili proposed a simple method based on tendency rate of equivalent micro-structures and showed such a method outperforms complex machine learning based methods [23]. It has been shown that the sign prediction problem can be considered as a matrix completion problem [10][12] [24]. Having such a formulation, different Matrix Factorization based techniques can be efficiently applied to this problem. These predictors first consider a sparse matrix corresponding to an input signed network. Then, the remaining entries are filled by employing matrix factorization strategy. In addition to the models based on Matrix Factorization and path based models, recently some algorithms based on probabilistic models as well as those based on embedding techniques have emerged to approach the problem [25-27]. Although the existing models for sign prediction have relatively good performance on homogenous networks, they have been originally designed for networks with only one type of relations, and cannot be directly applied to multilayer networks.

### 2.2. Link prediction in multilayer networks

Link prediction in networked structures is one of the heavily studied research topics in the field of social networks analysis and mining. The link prediction problem is often studied in an offline manner, that is by hiding a number of existing links, the structure of the network is used to predict their existence. Good link prediction algorithms are those that result in high probabilities for the hidden links. Different ink prediction algorithms can be categorized into two classes: those using local/global network statistics without learning and those based on machine learning methods. The first class of methods are those that take local properties of nodes, e.g. common neighbors, or global properties, e.g. path-based features such as Katz index, and obtain a probability of link existence without using any learning method. On the other hand, in the second class, first a number of features are extracted based on local and global nodal properties, and then a machine learning algorithm is used to solve a classification task with two classes: existence or non-existence of links. The link prediction algorithms have been recently considered for multilayer networks [28, 29]. Recently, Jalili et al. proposed a method based on features extracted from meta-paths and showed that it can successfully predict the missing links in multilayer structures [29]. They also showed that by employing information on multiple layers one can always improve performance of the link prediction tasks. Out proposed method in this manuscript is also based on meta-paths. We provide background information on meta-paths in the following.

### 2.3. Meta-Paths

As explained, the paths between two objects provide a good description about what is happening between them. Path-based features have been frequently applied to the problem of link prediction in homogeneous networks [30]. The same idea has been extended for the link prediction problem in heterogeneous networks [31, 32]. In a heterogeneous network, two objects can be connected by paths that cross different layers of the network or include different types of objects. These objects can be heavily connected to one another via different paths. Since heterogeneous networks include objects and relations of multiple types, employing the features originally proposed for homogeneous networks may result in information loss. In heterogeneous networks, two paths with different structures may indicate completely different type of relations. To account for these diverse types of connections among users, a methodology called meta-paths has been introduced [33].

*Meta-path* is a path defined on the network scheme, in which nodes represent object types, and relations of paths delineate types of connections. Let us first define a heterogeneous network. A *Heterogeneous network* is a network which contains multiple types of links and nodes, and can be represented as $G(V,E,T_v,T_E)$ with a link-type mapping function $\omega: E \rightarrow T_E$, where each link $e \in E$ belongs to a particular link type $\omega(e) \in T_E = \{t_{e,1}, t_{e,2}, \ldots, t_{e,n}\}$ and object-type mapping function $\tau: V \rightarrow T_V$ that maps each object $v \in V$ to a particular object type $\tau(v) \in T_V = \{t_{v,1}, t_{v,2}, \ldots, t_{v,n}\}$. In this paper, we denote networks that contain one type of objects and multi-types of relations, a special case of heterogeneous networks, often called multilayer or multiplex networks in the literature. In this case, each layer of network $G_i(V,E_i)$ includes edges belonging to a particular type of edge $t_{v,i}$ where $E_i = \{e | \omega(e) = t_{e,i}\}$.

Having the above definition for heterogeneous networks, a meta-path over graph $G(V,E,T_v,T_e)$ is denoted as $T_{V,l} \xrightarrow{T_{E,l}} T_{V,m} \xrightarrow{T_{E,m}} \ldots \xrightarrow{T_{E,n}} T_{V,n}$ and the length of the path is defined as the number of links involved in it. For example, in Facebook network, given two users $u_a$ and $u_b$, who both liked an image, say $I_v$, the relation can be denoted as a path $u_a \xrightarrow{likes} I_v \xrightarrow{likes^{-1}} u_b$. This path follows the meta-path $U \xrightarrow{L} I \xrightarrow{L^{-1}} U$, in which $U$ represents the type user, $I$ denotes the object type that is image in this example, and $L$ represents the *Like* relationship between a user and an image. Also, $I \xrightarrow{L^{-1}} U$ means $U \xrightarrow{L} I$. Another example can be $u_a \xrightarrow{follows} u_b \xrightarrow{friendship} u_c$, which describes the relationship that $u_a$ follows $u_b$ and $u_b$ is friend of $u_c$. The topology of the path can be described as $U \xrightarrow{Follow} U \xrightarrow{Friend} U$. In these examples, a path has length two. To define meta-path based features, first, the type of meta-paths between two target objects should be extracted up to a predefined path length based on traversing the target network scheme. This can be obtained using standard traversal methods such as the Breadth First Search (BFS). Once the meta-paths are determined, the next stage is to propose measures on these meta-paths. Various measures have been introduced on meta-paths [32]. Here, we employ those based on path count, which measure the number of path instances between two objects following a given meta-path.

## 3. Prediction of feedback on link initiation

### 2.1. Dataset

Because of privacy issues, the friendship requests, and the positive/negative responses are not visible to the public in any social networking platform, and thus the two-way relations cannot be gathered from well-known social networks like Linkedin and Facebook. Indeed, in such networks, we can only extract the network of friendships between the users. Considering these limitations, in this paper, we introduce a dataset that has been extracted from a social gaming website called MasterCup. The user-blind data has been provided by the website's admin for our research. The MsterCup dataset belong to a popular local sport website (varzesh3.com), which ranks 346 worldwide according to Alexa (accessed Feb 2018). In the MasterCup environment, each user has the role of a team coach for a soccer simulation team and can modify their team's strategy and choose a framework for it. The framework consists of multiple leagues in different levels in which the teams compete. When users join the framework, first they start in the most amateur league, and then progress to higher-level leagues based on the performance. Besides the regular matches in each league, the users can invite others to friendly matches. Such matches will only be held if the invitation is accepted by a user. On top of these two types of interactions, the users can send or receive messages. Therefore, the relations among users in this framework can be classified into three classes: regular matches, friendly matches, and message interactions. The friendly matches can be classified as two-way relations and the other two types as one-way relations. In the following, we describe the heterogeneous network obtained from the dataset.

Based on the above definition, the MasterCup dataset can be represented as a heterogeneous network. The dataset includes one type of object, users shown as $T_v=\{U\}$, and three types of interactions, $T_e=\{F, R, M\}$. As an abbreviation, initial capital letters are used to denote the connections, namely *F* for friendly matches, *R* for regular matches and *M* for message interactions. We represent layers of the network for each type of relations *F*, *R* and *M* as $G_F$, $G_R$ and $G_M$, respectively. Table 1 shows the statistics for the dataset. Heterogeneous networks can be represented using a meta-structure called network scheme. In the network scheme, nodes are types of objects and the edges between the nodes are relations between types. Figure 1 demonstrates the network schema for the heterogeneous MasterCup network. The interactions for regular matches are one way. To model these interactions, consistent with the status theory, we consider a match between two users as a directed edge, in which the direction is from the loser (lower status) to the winner (higher status). Also, the message interaction is modeled as a directed network from the sending users to the receiving ones.

Modeling the two-way relationships should be made in such a way that minimal information is lost. For example, if one models a two-way relation from *u* to *v* as a directed edge *e(u,v)* and the positive response as a directed edge for *v* to *u*, *e(v,u)*, an asymmetric edge *e(u,v)* can be interpreted as the negative response to the link initiation from *u* to *v*. However, such a model leads to information loss to some extent. Suppose that both links *e(u,v)* and *e(v,u)* exist in the network. In this case, the responding and initiating nodes are not distinguishable. The problem can be solved by adding labels on the edge. Although the labeled network keeps the whole information, it has a relatively high complexity which makes analysis of the network difficult. To overcome such problems, here we represent two-way interactions as directed signed networks. A *Signed network* is a network with a sign on all its edges, which means that for edge *e(u,v)* from node *u* to node *v*, we have a label denoted as *s(u,v)*, which can take {+1,-1} values. That is, we represent link request from node *u* to node *v* as a directed link from *u* to *v* and the response given to this link initiation as a sign on this edge. The responses to link initiations have two classes: positive (acceptance) and negative (rejections) and they can be mapped to positive and negative labels, respectively.

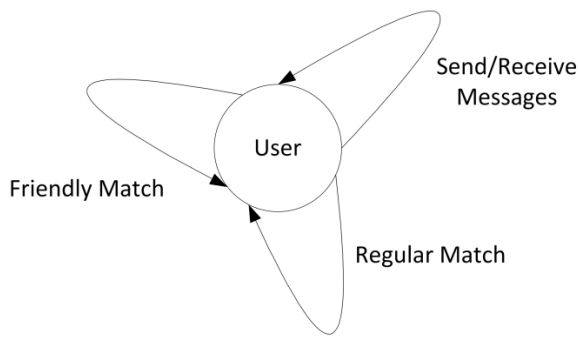

Figure 1: Scheme for MasterCup network

Table 1: Statistics of the heterogeneous MasterCup network ($G_F$, $G_M$ and $G_R$ are different layers of the network)

| Number of users | Number of edges on $G_F$ | Number of edges on $G_M$ | Number of edges on $G_R$ |
|---|---|---|---|
| 44124 | 182598 | 1354606 | 1448620 |

## 3.2. Response prediction for link initiation as the problem of edge sign prediction

Considering our model for a network of two-way relations, the response prediction task can be considered as the problem of edge sign prediction in signed networks. The sign prediction problem is defined as follows. Suppose that we are given a signed network in which the sign for one of the target edges (the edge from an initiating node to a recipient node) is not available. The sign prediction task is defined as how the hidden label can be inferred based on the information obtained from the rest of the network. We consider $G_F$ as the target layer and the other two layers ($G_R$ and $G_M$) as source layers and show that information obtained from the source layers can be useful in predicting the signs in the target layer. We introduce a method to incorporate the information obtained from different layers of the network to predict the signs in the layer that involve the two-way relations. Our method is based on nodes-based and cluster-based features obtained from meta-paths, which are then used in a machine learning task to solve the classification problem with two classes, positive or negative signs for edges, which indeed represent acceptance or rejection of the invitation. The details of these features are in the following.

### 3.2.1. Node-based meta-path features for sign prediction

Let us consider two users $u_a$ an $u_b$ for which the sign of the link from $u_a$ to $u_b$ is to be predicted. The node-based meta-paths defined in this work has the form $U \xrightarrow{X} U \xrightarrow{F} U$, where $X$ represents one-way relations in $R$ and $M$ layers, and $F$ represent the positive/negative signs modeled for two-way relations. Indeed in this way, we assume that the sign of the edge between user $u_b$ and those related to $u_a$ are informative about the sign of the edge from $u_a$ to $u_b$. Apparently, the simplest paths involving multi-type links are the paths with length two. By traversing the MasterCup network, the paths with multi-type connections of the form $U \xrightarrow{X} U \xrightarrow{F} U$ are represented in table 2. As it can be seen, the relationships of type $R$ and $M$ have been used to define users that are related to the target initiating user. The links that have type $F$ can take two labels: positive or negative. We define a meta-path to distinguish these values. In other words, we use source layers ($G_M$ and $G_R$) to define users that are related to the target initiating user. We denote the predictor obtained from features of node-based meta-paths as NB-MP predictor. To extract features based on path counts, we should take into account the number of the paths following the defined meta-paths. One can define features based on complex topological structures, which may better describe relationships between the target nodes. However, it might reduce reliability of the information obtained from the features.

Table 2: List of node-based meta-paths with length two crossing different layers of the MasterCup network.

| List of meta-paths that has the form $U \xrightarrow{X} U \xrightarrow{F} U$ | |
|---|---|
| $U \xrightarrow{R} U \xrightarrow{F^{-1}(-)} U$ | $U \xrightarrow{R} U \xrightarrow{F(-)} U$ |
| $U \xrightarrow{R} U \xrightarrow{F^{-1}(+)} U$ | $U \xrightarrow{R} U \xrightarrow{F(+)} U$ |
| $U \xrightarrow{R^{-1}} U \xrightarrow{F^{-1}(-)} U$ | $U \xrightarrow{R^{-1}} U \xrightarrow{F(-)} U$ |
| $U \xrightarrow{R^{-1}} U \xrightarrow{F^{-1}(+)} U$ | $U \xrightarrow{R^{-1}} U \xrightarrow{F(+)} U$ |
| $U \xrightarrow{M} U \xrightarrow{F^{-1}(-)} U$ | $U \xrightarrow{M} U \xrightarrow{F(-)} U$ |
| $U \xrightarrow{M} U \xrightarrow{F^{-1}(+)} U$ | $U \xrightarrow{M} U \xrightarrow{F(+)} U$ |
| $U \xrightarrow{M^{-1}} U \xrightarrow{F^{-1}(-)} U$ | $U \xrightarrow{M^{-1}} U \xrightarrow{F(-)} U$ |
| $U \xrightarrow{M^{-1}} U \xrightarrow{F^{-1}(+)} U$ | $U \xrightarrow{M^{-1}} U \xrightarrow{F(+)} U$ |

### 3.2.2. Cluster-based meta-paths for sign-prediction

In general, it can be argued that in order to build an effective prediction model based on meta-paths one of the following conditions should be fulfilled, *i*) the network should be extremely dense, which is not the case in many real-world networks, or *ii*) the network should have a hierarchical structure. In a multilayer network, in which all objects are of the same type, we do not have an object that could play the role of bridge in connecting the target objects. Similarly, in multilayer MasterCup network, users are the only objects. Moreover, the target signed network is extremely sparse in this dataset. Therefore, if we define meta-paths based on the objects, we will not have enough number of paths for each meta-path, which results in an inaccurate prediction model. Obviously, density of the input network is a factor that cannot be changed, and thus the structure of the input network should be modified on multi-layer networks. We aim at adding new objects to the input network which have the role of bridge in connecting the target nodes, thus to obtain a hierarchical structure for the network. Considering the multilayer structure, we first cluster each layer of the network separately. Here, we use InfoMap algorithm [34], which is one of the well-known community detection algorithms. Then, we define new type of objects based on the obtained clusters. Figure 2 is a toy example describing how we add the cluster-based objects to the input network. As it can be seen, in this example the input network has three clusters, $c_1$, $c_2$ and $c_3$. Each cluster is modeled as a node of type cluster and added to the network.

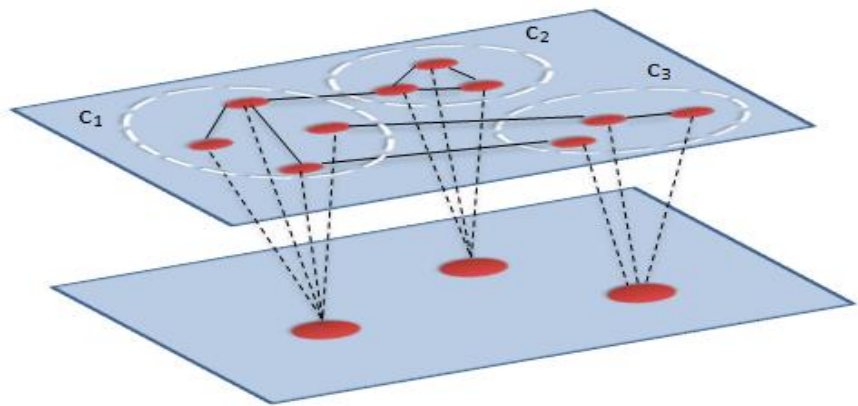

Figure 2: A sample network with three clusters (upper layer) and the nodes representing each cluster (lower layer)

Here, we first separately cluster the source networks $G_R$ and $G_M$, where the set of clusters for $G_R$ and $G_M$ are represented by $C_R$ and $C_M$, respectively. These clusters are then added to the network as two new types of objects. Indeed, we model each cluster as a node where its type represents the set of clusters it belongs. Consequently, we add a new type of relationship, namely *belong*, denoted by $B(U \xrightarrow{B} C)$ between users and clusters. The network is then represented as $G'(V',E', T_{v'}, T_{E'})$ with $T_{v'}=\{U, C_R, C_M\}$ and $T_{v'}=\{F,R,M,B\}$. Figure 3 represents the scheme of the MasterCup network after including the cluster-based objects.

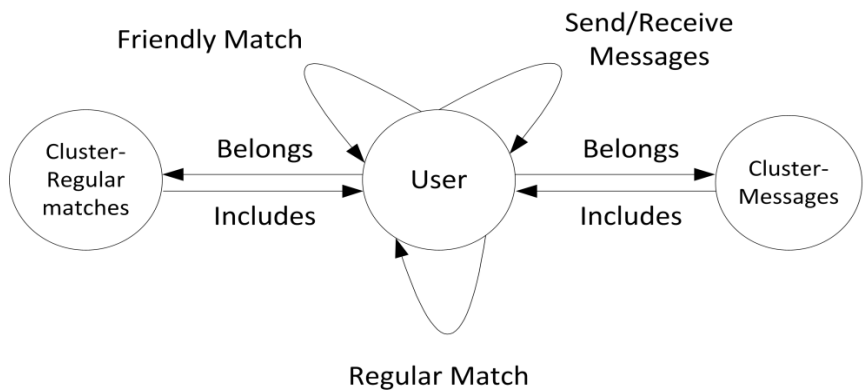

Figure 3: Scheme of the MasterCup network after including the Cluster-based objects.

Having these cluster-based objects in the network scheme of the MasterCup network, we define meta-paths including the objects of type cluster. We denote this form of meta-paths as cluster-based meta-paths. Table 3 represents the list of meta-paths we use to obtain cluster-based meta-paths with length 4. As mentioned, the meta-paths we define for the edge sign prediction has the form: $U \xrightarrow{X} U \xrightarrow{F} U$. In node-based features $X$ is a relationship with length 1, which can take relations of type $M$ and $R$. Having cluster-based objects, here we define $X$ as a composite relationship with length 3 that has the form $\xrightarrow{X'} U \xrightarrow{B} C \xrightarrow{B^{-1}}$ and $X'$ takes relationships of type $M$, and $R$ and $C$ can take object types $C_R$ and $C_M$, respectively. Based on the form of meta-paths as $U \xrightarrow{X'} U \xrightarrow{B} C \xrightarrow{B^{-1}} U \xrightarrow{F} U$, 16 meta-paths can be obtained (table 5). It is expected that the number of paths following cluster-based meta-path to be much higher than those following the node-based meta-paths. We denote the predictor obtained from cluster-based meta-paths as CB-MP based predictor.

Table 3: List of cluster-based meta-paths with length four over the MasterCup network.

| List of the cluster-based meta-paths that has the form $U \xrightarrow{X'} U \xrightarrow{B} C \xrightarrow{B^{-1}} U \xrightarrow{F} U$ | |
|---|---|
| $U \xrightarrow{R} U \xrightarrow{B} C_R \xrightarrow{B^{-1}} U \xrightarrow{F(+)} U$ | $U \xrightarrow{R} U \xrightarrow{B} C_R \xrightarrow{B^{-1}} U \xrightarrow{F^{-1}(+)} U$ |
| $U \xrightarrow{R} U \xrightarrow{B} C_R \xrightarrow{B^{-1}} U \xrightarrow{F(-)} U$ | $U \xrightarrow{R} U \xrightarrow{B} C_R \xrightarrow{B^{-1}} U \xrightarrow{F^{-1}(-)} U$ |
| $U \xrightarrow{R^{-1}} U \xrightarrow{B} C_R \xrightarrow{B^{-1}} U \xrightarrow{F(+)} U$ | $U \xrightarrow{R^{-1}} U \xrightarrow{B} C_R \xrightarrow{B^{-1}} U \xrightarrow{F^{-1}(+)} U$ |
| $U \xrightarrow{R^{-1}} U \xrightarrow{B} C_R \xrightarrow{B^{-1}} U \xrightarrow{F(-)} U$ | $U \xrightarrow{R^{-1}} U \xrightarrow{B} C_R \xrightarrow{B^{-1}} U \xrightarrow{F^{-1}(-)} U$ |
| $U \xrightarrow{M} U \xrightarrow{B} C_M \xrightarrow{B^{-1}} U \xrightarrow{F(+)} U$ | $U \xrightarrow{M} U \xrightarrow{B} C_M \xrightarrow{B^{-1}} U \xrightarrow{F^{-1}(+)} U$ |
| $U \xrightarrow{M} U \xrightarrow{B} C_M \xrightarrow{B^{-1}} U \xrightarrow{F(-)} U$ | $U \xrightarrow{M} U \xrightarrow{B} C_M \xrightarrow{B^{-1}} U \xrightarrow{F^{-1}(-)} U$ |
| $U \xrightarrow{M^{-1}} U \xrightarrow{B} C_M \xrightarrow{B^{-1}} U \xrightarrow{F(+)} U$ | $U \xrightarrow{M^{-1}} U \xrightarrow{B} C_M \xrightarrow{B^{-1}} U \xrightarrow{F^{-1}(+)} U$ |
| $U \xrightarrow{M^{-1}} U \xrightarrow{B} C_M \xrightarrow{B^{-1}} U \xrightarrow{F(-)} U$ | $U \xrightarrow{M^{-1}} U \xrightarrow{B} C_M \xrightarrow{B^{-1}} U \xrightarrow{F^{-1}(-)} U$ |

Although, we proposed the cluster-based meta-paths to obtain a hierarchical structure to deal with the sparsity problem, the model can also be justified from other aspects. CB-MP can be interpreted as NB-MP with a varying length. One of the main issues in defining the node-based meta-paths is how to set the length of the meta-paths [33]. As mentioned, in order to predict the sign of the edge from $u$ to $v$, we define features of type $U \xrightarrow{X} U \xrightarrow{F} U$, such that $X$ can take any type of composite relationships between the target initiator and its neighbors. However, the question is how to define these paths and determine their optimal length. The distance between two nodes might not be a precise factor for determining strength of their relationship, and the structure of the network can be a better indicator. Therefore, one can employ the structure of the networks to discriminate the related objects to the target node. From this point of view, the cluster-based meta-paths approaches to this problem by defining length of the node-based meta-paths.

## 4. Results and discussion

In this section, we provide the experiments in order to assess the effectiveness of the proposed model. Before presenting the main results, let us first study properties of the MasterCup network and to further justify the response prediction problem and the intuition behind the proposed approach.

### *4.1. layer-layer correlation of MasterCup network*

In this experiment, we compare the structure of different layers of the MasterCup network. Such analysis can give some clue on how to design a model to extract and transform information from one layer to do predictions on the other one. Here, Kendall's $\tau$-rank correlation coefficient is used to analyse inter-layer degree correlation, normalized hamming distance is employed to measure the inter-layer correlation of node activities and the number of common edges is used to study the inter-layer correlation of edge presence. Despite these metrics are coarse-grained, they can present an overall summary of the correlations of important structural patterns in the multiplex network.

Kendall's $\tau$-rank correlation coefficient takes a value between -1 and 1 and measures the degree correlation between two layers. Table 4 shows Kendall's $\tau$ correlation coefficient between each pair of layers. In order to have more meaningful results, isolated nodes of each layer are eliminated from both layers. It is seen that the correlation between $G_M$ and $G_F$ layers as well as $G_R$ and $G_F$ layers is small, while there is significant degree correlation between $G_M$ and $G_R$ layers, indicating the fact that the nodes that have more interactions with others also send and receive more messages. It is also notable that in-degree correlation between $G_M$ and $G_F$ layers is more than the out-degree correlation, while this relation is reversed for $G_R$ and $G_F$ layers.

Normalized Hamming distance is used to analyze the correlation of node activity in two layers. It takes a value between 0 and 1, where 0 indicates that all active nodes in one layer are also active in the other one and 1 implies that the activities of the nodes are completely independent in the two layers. Here, a node is considered in-active (out-active) for the cases when its in-degree (out-degree) is above 0. As it can be seen in Table 1, the lowest correlation is out-activity correlation of $G_M$ and $G_R$ layers and the highest one is in-activity correlation of $G_F$ and $G_M$ layers. The number of common edges counts the edges existing in a pair of layers. Taking only $G_F$ layer into account, about a quarter of its edges are also present in $G_M$ layer and 10820 edges of this layer are available in either $G_M$ or $G_R$ layers. Moreover, the results show that only 1250 edges are common in all the layers. Interestingly, more than 75% of edges of $G_F$ layer do not exist in the other layers, indicating that these relations are formed based on interactions that are not directly available in other layers. Based on these results, we can conclude that the structure of the layers are not highly correlated. Therefore, to incorporate information from source networks to predict the links in a target network, relying on correlation between the source and target layers might not result in good outcome. Indeed, the patterns of relations between links of two different layers are more complex than co-occurrence of link between two objects in different layers.

Table 4: Correlations of different layers ($G_M$, $G_R$ and $G_F$) of the MasterCup network.

| Pair of layers | Kendall's $\tau$ correlation coefficient | | Hamming Distance | | Number of common edges |
|---|---|---|---|---|---|
| | In-degree | Out-degree | In-activity | Out-activity | |
| $G_F - G_R$ | 0.164 | 0.218 | 0.543 | 0.485 | 9358 |
| $G_F - G_M$ | 0.315 | 0.225 | 0.820 | 0.655 | 2712 |
| $G_M - G_R$ | 0.610 | 0.626 | 0.584 | 0.338 | 224356 |

## *4.2. Acceptance rate of a link initiation as a function of embededness*

Most of recommender systems implemented on social networks work based on collaborative filtering algorithms, that use a similarity metric such as embeddedness that is common neighbors between users [35]. In this experiment, we investigate effectiveness of such policy for a user-user recommender system. Considering MasterCup dataset, a recommender system which wants to recommend some other users to the target user for friendly matches, should perform this based on the number of users who have played friendly matches with both the target and candidate users. Here, we aim at verifying whether there is any dependency between the number of common neighbors and the type of responses given to the link initiation. Figure 4 represents the percentage of link initiations (which received positive/negative responses) as a function of embededness. As it can be seen, almost 85% of the link initiations which received negative responses have zero embededness. Moreover, as the embededness increases, the probability of link initiations with a positive response increases. These results indicate that the users with more mutual friends with the target user are more likely to give positive response to the link initiation made by the target user.

Regarding these results, there are two issues that should be considered. First, a majority of link initiations (around 55%) which received positive response have zero embededness. This means that a large group of recommendation options which may give positive feedback to the target user's link initiation would be eliminated by a classic recommender system that are based on common neighbors. Also, we may say that this group of users is more unexpected recommendations than those that have large number of common neighbors with the target user. Indeed, the target user is likely to know about those with high embededness. Therefore, filling recommendation lists with such users decreases information of recommendations, since it lowers unexpectedness of recommendations. This issue somehow justifies employing different type of interactions to answer the link initiation feedback problem. The second issue is that a considerable proportion of link initiations with high embededness received negative response. For example, about 25% of initiations with embededness of 4 received negative feedback. It means that a recommender system relying only on common neighbors may result in small prediction accuracy.

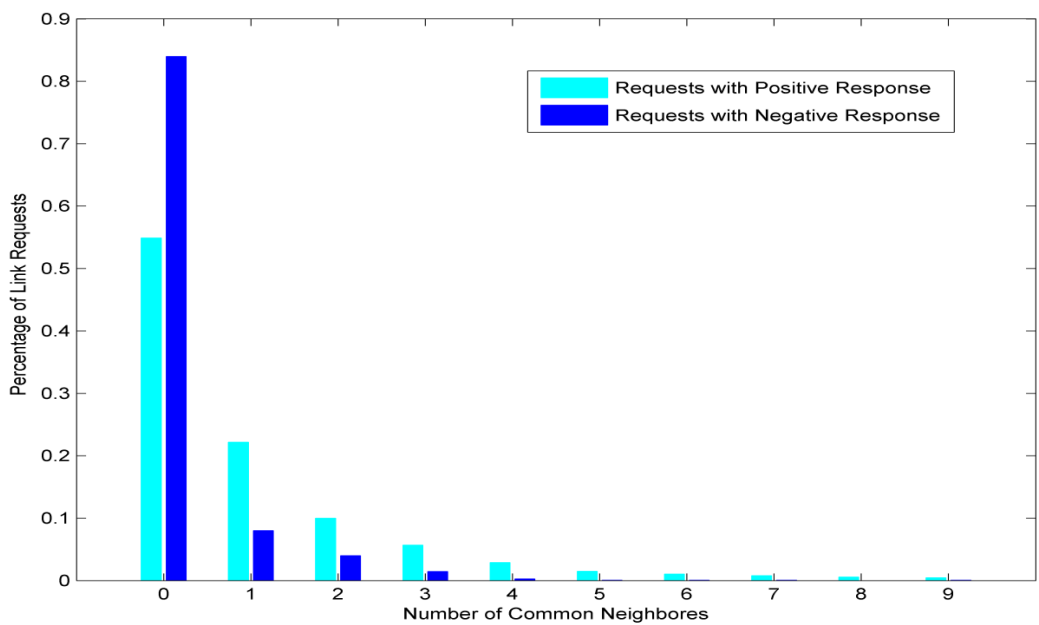

Figure 4: Percentage of link initiations which receive negative or positive responses as a function of the number of common neighbors between the initiator node and recipient node.

### 4.3. *Performance of link initiation feedback predictors*
Our experiments are based on evaluating different prediction algorithms using 10-fold cross validation, where in each fold, 10% of the original dataset is considered as test set and 90% as training set. For each experiment, we first learn the model based on the training set, and then perform the sign prediction task on the test set. The message and interaction networks ($G_M$ and $G_R$) are the same for all the experiments. Accuracy of classification is a frequently used metrics to evaluate the quality of predictions. However, in networks of two-way relations, the number of positive edges is considerably higher than negative ones, and comparing prediction methods based on the accuracy of the original test sets can be misleading in some cases. Thus, we use balanced accuracy to avoid such misleading results [9]. In this method, the experiments are performed on the original test and training datasets, and then the mean true positive rate of the predictions is reported as balanced accuracy. It is worth to mention that in the learning phase of the proposed model, we employ cost sensitive learning model as the number of training samples with positive and negative labels are not the same.

In this experiment, we compare the performance of the proposed prediction model (CB-MP based predictor) with three other predictors: NB-MP predictor and two models originally proposed for homogeneous networks (machine learning based model using a set of path-based features introduced by Leskovec et al. (dented as NB-SN model) and a model based on Matrix Factorization (MF) introduced in [10]). For both CB-MP and NB-MP, we use Support Vector Machines (SVM) as the learning algorithm. The set of meta-paths used for CB-MP and NB-MP predictors are the meta-paths which are introduced in the Tables 3 and 2, respectively. As it can be seen, the number of the meta-paths (features) used in NB-MP is the same as the CB-MP. Therefore, the learning phase of building prediction models for both predictors has the same computational complexity.

Among the prediction models, three of them including CB-MP, NB-MP and NB-SP, are based on path-based features, while MF is based on a quite different methodology which employs Matrix Factorization for the prediction task. The complexity of MF is higher than the other three. Figure 5 shows the accuracy of predictions; it can be seen that CB-MP and NB-MP outperforms NB-SP, which is designed for homogeneous networks. This is mainly due to the fact that in both CB-MP and NB-MP models we incorporate more information resources in building the prediction model. CB-MP significantly outperforms the MF model, which itself has been shown to be more efficient than traditional features-based models on homogeneous networks [10]. These results indicate that the information obtained from source layers is not negligible, although source layers are not highly correlated with the target layer. Clearly, all methods outperform the Random predictor, which has accuracy of 50%. The outperformance of CB-MP over NB-MP confirms our assumption in the cluster-based met-paths. One may argue that CB-MP predictor is a stronger predictor (compared to NB-MP) as it considers the sparsity issue of the multilayer network. From another point of view, the features obtained in CB-MP model are much more reliable than those obtained in NB-MP model. Such higher reliability values can be linked to the hierarchical structure of the network used in CB-MP predictor.

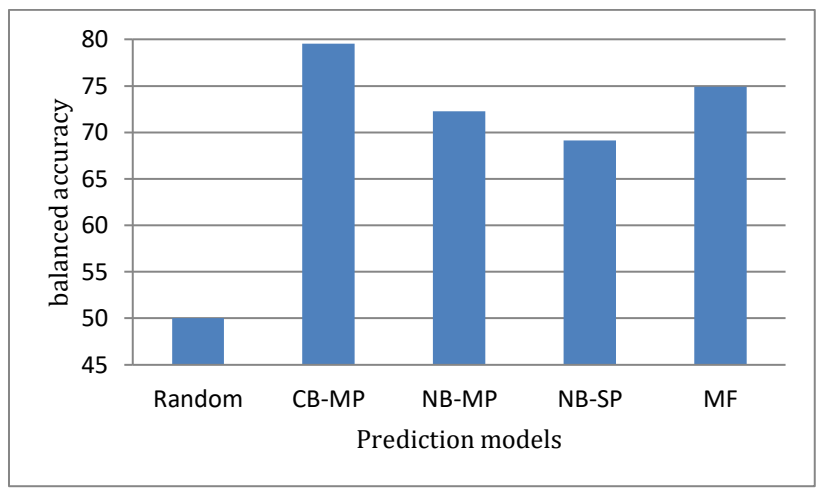

Figure 5: Accuracy of five prediction models: Cluster based Meta-Path (CB-MP), Node based Meta-Path (NB-MP), path-based model on homogeneous network (NB-SP), Matrix Factorization (MF) and Random predictors for edge sign prediction task on MasterCup network.

## 5. Conclusion
In this paper, we introduced a new type of relationships in online social networks denoted as two-way relationships; a type of interaction which is created based on a request and a subsequent (positive or negative) response. Such relationships can be frequently found in many online social networks. We studied the response prediction problem over the network of two-way relationships. The problem is to predict the (positive or negative) response of users when they receive requests from other users. As a case study, we focused on a dataset obtained from a social gaming website, and modeled it as a three-layer network. In order to predict the response of users, not only we employed connectivity network of two-way relationships, but also other type of interactions between the users based on a meta-path based methodology. Moreover, we extended the methodology and proposed cluster-based meta-paths to make our prediction model more reliable on sparse networks, which is often the case for many real networks. The effectiveness of the proposed prediction model was evaluated on the dataset, and the results revealed outperformance of the proposed cluster-based method over other methods. Our proposed approach can be used to create link recommendation systems for social networks.

## Acknowledgement

Mahdi Jalili is supported by Australian Research Council through project No. DP170102303.

## References

[1] M. Kivelä, A. Arenas, M. Barthelemy, J. P. Gleeson, Y. Moreno, and M. A. Porter, "Multilayer networks," *arXiv preprint arXiv:1309.7233,* 2013.
[2] S. Boccaletti, G. Bianconi, R. Criado, C. Del Genio, J. Gómez-Gardeñes, M. Romance*, et al.*, "The structure and dynamics of multilayer networks," *Physics Reports,* 2014.
[3] S.-H. Yang, B. Long, A. Smola, N. Sadagopan, Z. Zheng, and H. Zha, "Like like alike: joint friendship and interest propagation in social networks," in *Proceedings of the 20th international conference on World wide web*, 2011, pp. 537-546.
[4] !!! INVALID CITATION !!! [4-6].
[5] J. Chen, W. Geyer, C. Dugan, M. Muller, and I. Guy, "Make new friends, but keep the old: recommending people on social networking sites," in *Proceedings of the SIGCHI Conference on Human Factors in Computing Systems*, 2009, pp. 201-210.
[6] P. Moradi, F. Rezaeimehr, S. Ahmadian, and M. Jalili, "TCARS: time- and community-aware recommendation system," *Future Generation Computer Systems,* vol. 78, pp. 419-429, 2018.
[7] M. Ranjbar, P. Moradi, M. Azami, and M. Jalili, "An imputation-based matrix factorization method for improving accuracy of collaborative filtering systems," *Engineering Applications of Artificial Intelligence,* vol. 46, pp. 58-66, 2015.
[8] J. Leskovec, D. Huttenlocher, and J. Kleinberg, "Signed networks in social media," in *Proceedings of the SIGCHI Conference on Human Factors in Computing Systems*, 2010, pp. 1361-1370.
[9] A. Javari and M. Jalili, "Cluster-Based Collaborative Filtering for Sign Prediction in Social Networks with Positive and Negative Links," *ACM Transactions on Intelligent Systems and Technology (TIST),* vol. 5, p. 24, 2014.
[10] C.-J. Hsieh, K.-Y. Chiang, and I. S. Dhillon, "Low rank modeling of signed networks," in *Proceedings of the 18th ACM SIGKDD international conference on Knowledge discovery and data mining*, 2012, pp. 507-515.
[11] J. Leskovec, D. Huttenlocher, and J. Kleinberg, "Predicting positive and negative links in online social networks," in *Proceedings of the 19th international conference on World wide web*, 2010, pp. 641-650.
[12] T. DuBois, J. Golbeck, and A. Srinivasan, "Predicting trust and distrust in social networks," in *Privacy, security, risk and trust (passat), 2011 ieee third international conference on and 2011 ieee third international conference on social computing (socialcom)*, 2011, pp. 418-424.

[13]	P. Doreian and A. Mrvar, "Partitioning signed social networks," *Social Networks,* vol. 31, pp. 1-11, 2009.
[14]	P. Esmailian, S. E. Abtahi, and M. Jalili, "Mesoscopic analysis of online social networks: The role of negative ties," *Physical review E,* vol. 90, p. 042817, 2014.
[15]	M. Kivelä, A. Arenas, M. Barthelemy, J. P. Gleeson, Y. Moreno, and M. A. Porter, "Multilayer networks," *Journal of Complex Networks,* vol. 2, pp. 203-271, 2014.
[16]	J. Tang, Y. Chang, C. Aggarwal, and H. Liu, "A survey of signed network mining in social media," *ACM Computing Surveys (CSUR),* vol. 49, p. 42, 2016.
[17]	G. Bachi, M. Coscia, A. Monreale, and F. Giannotti, "Classifying trust/distrust relationships in online social networks," in *Privacy, Security, Risk and Trust (PASSAT), 2012 International Conference on and 2012 International Confernece on Social Computing (SocialCom)*, 2012, pp. 552-557.
[18]	J. Leskovec, D. Huttenlocher, and J. Kleinberg, "Predicting positive and negative links in online social networks," in *Proceedings of the 19th international conference on World wide web*, 2010, pp. 641-650
[19]	R. Guha, R. Kumar, P. Raghavan, and A. Tomkins, "Propagation of trust and distrust," in *Proceedings of the 13th international conference on World Wide Web*, 2004, pp. 403 - 412
[20]	K.-Y. Chiang, N. Natarajan, A. Tewari, and I. S. Dhillon, "Exploiting longer cycles for link prediction in signed networks," in *Proceedings of the 20th ACM international conference on Information and knowledge management*, 2011, pp. 1157-1162.
[21]	M. Shahriari, O. Askari Sichani, J. Gharibshah, and M. Jalili, "Sign prediction in social networks based on users reputation and optimism Authors," *Social Network Analysis and Mining,* vol. 6, p. 91, 2016.
[22]	M. Shahriari and M. Jalili, "Ranking nodes in signed social networks," *Social Network Analysis and Mining,* vol. 4, pp. 1-12, 2014.
[23]	A. Khodadadi and M. Jalili, "Sign prediction in social networks based on tendency rate of equivalent micro-structures," *Neurocomputing,* vol. 257, pp. 175-184, 2017.
[24]	P. Agrawal, V. K. Garg, and R. Narayanam, "Link Label Prediction in Signed Social Networks," in *IJCAI*, 2013, pp. 2591-2597.
[25]	M. R. Islam, B. A. Prakash, and N. Ramakrishnan, "SIGNet: Scalable Embeddings for Signed Networks," *arXiv preprint arXiv:1702.06819,* 2017.
[26]	S. Wang, J. Tang, C. Aggarwal, Y. Chang, and H. Liu, "Signed network embedding in social media," in *Proceedings of the 2017 SIAM International Conference on Data Mining*, 2017, pp. 327-335.
[27]	A. Javari, H. Qiu, E. Barzegaran, M. Jalili, and K. C.-C. Chang, "Statistical Link Label Modeling for Sign Prediction: Smoothing Sparsity by Joining Local and Global Information," in *2017 IEEE International Conference on Data Mining (ICDM)*, 2017, pp. 1039-1044.
[28]	D. Hristova, A. Noulas, C. Brown, M. Musolesi, and C. Mascolo, "A multilayer approach to multiplexity and link prediction in online geo-social networks," *EPJ Data Science,* vol. 5, p. 24, 2016.
[29]	M. Jalili, Y. Orouskhani, M. Asgari, N. Alipourfard, and M. Perc, "Link prediction in multilayer online social networks," *Royal Society Open Science,* vol. 4, p. 160863, 2017.
[30]	L. Lü and T. Zhou, "Link prediction in complex networks: A survey," *Physica A: Statistical Mechanics and its Applications,* vol. 390, pp. 1150-1170, 2011.
[31]	Y. Sun, R. Barber, M. Gupta, C. C. Aggarwal, and J. Han, "Co-author relationship prediction in heterogeneous bibliographic networks," in *Advances in Social Networks Analysis and Mining (ASONAM), 2011 International Conference on*, 2011, pp. 121-128.
[32]	Y. Sun and J. Han, "Mining heterogeneous information networks: principles and methodologies," *Synthesis Lectures on Data Mining and Knowledge Discovery,* vol. 3, pp. 1-159, 2012.

[33]   Y. Sun and J. Han, "Mining heterogeneous information networks: a structural analysis approach," *ACM SIGKDD Explorations Newsletter,* vol. 14, pp. 20-28, 2013.
[34]   A. Lancichinetti and S. Fortunato, "Community detection algorithms: a comparative analysis," *Physical review E,* vol. 80, p. 056117, 2009.
[35]   J. Bobadilla, F. Ortega, A. Hernando, and A. Gutiérrez, "Recommender systems survey," *Knowledge-based systems,* vol. 46, pp. 109-132, 2013.